\documentclass[11pt]{article}
\usepackage[preprint]{acl}

\usepackage{times}
\usepackage{latexsym}
\usepackage{booktabs}
\usepackage{xcolor,colortbl}
\usepackage{graphicx}
\usepackage{multirow}
\usepackage{booktabs,threeparttable,xcolor,colortbl}
\usepackage{tcolorbox}
\tcbuselibrary{skins,breakable}
\usepackage{algorithm}
\usepackage{enumitem}
\usepackage{algorithmic}
\usepackage{array}
\usepackage{amssymb}
\usepackage{booktabs}
\usepackage{multirow}
\usepackage{enumitem}
\usepackage{graphicx}
\usepackage{amsmath}
\tcbuselibrary{listings, breakable}

\definecolor{headerblue}{HTML}{A2C8E9}
\definecolor{datacellgreen}{HTML}{9DDDA5}   
\definecolor{bestoverall}{HTML}{F8CC87}
\newcommand{\best}[1]{\cellcolor{datacellgreen}\textbf{#1}}         % best within a method block
\newcommand{\overallbest}[1]{\cellcolor{bestoverall}\textbf{#1}}  

\usepackage[T1]{fontenc}
\usepackage[utf8]{inputenc}

\usepackage{microtype}

\usepackage{inconsolata}

\usepackage{graphicx}

\tcbset{
  promptstyle/.style={
    enhanced,
    breakable,
    width=\linewidth,

    boxrule=0.45pt,
    arc=2pt,

    colframe=black!35,
    colback=black!2,
    coltitle=black,
    colbacktitle=black!8,

    left=2mm, right=2mm, top=1.5mm, bottom=1.5mm,
    boxsep=1mm,
    before skip=6pt,
    after skip=6pt,

    fonttitle=\bfseries\footnotesize,
    fontupper=\footnotesize\raggedright,

    attach boxed title to top left={xshift=2mm, yshift*=-1mm},
    boxed title style={
      colback=black!8,
      colframe=black!35,
      boxrule=0.45pt,
      arc=2pt,
      left=1mm, right=1mm, top=0.5mm, bottom=0.5mm
    },

    before upper={
      \setlength{\parindent}{0pt}
      \setlist[itemize]{nosep,leftmargin=1.2em,topsep=2pt,itemsep=1pt,parsep=0pt}
      \setlist[enumerate]{nosep,leftmargin=1.4em,topsep=2pt,itemsep=1pt,parsep=0pt}
    }
  }
}

\title{LLM-as-RNN: A Recurrent Language Model for Memory Updates and Sequence Prediction}
\newcommand{\affA}{\textsuperscript{1}}
\newcommand{\affB}{\textsuperscript{2}}
\newcommand{\affC}{\textsuperscript{3}}
\newcommand{\affD}{\textsuperscript{4}}
\newcommand{\affE}{\textsuperscript{5}}

\author{
  Yuxing Lu\thanks{Equal contribution.}\affA\affB,
  J. Ben Tamo\footnotemark[1]\affA,
  Weichen Zhao\affC,
  Nan Sun\affD,
  Yishan Zhong\affA,
  Wenqi Shi\affE,
  \\
  \textbf{Jinzhuo Wang\thanks{Corresponding author.}\affB,}
  \textbf{May D. Wang\footnotemark[2]\affA}
  \\[6pt]
  \textsuperscript{1}Georgia Institute of Technology,
  \textsuperscript{2}Peking University,
  \textsuperscript{3}Shandong University,
  \\
  \textsuperscript{4}Huazhong University of Science and Technology,
  \textsuperscript{5}UT Southwestern Medical Center
}

\begin{document}
\maketitle
\begin{abstract}
Large language models are strong sequence predictors, yet standard inference relies on immutable context histories. After making an error at generation step $t$, the model lacks an updatable memory mechanism that improves predictions for step $t{+}1$. We propose LLM-as-RNN, an inference-only framework that turns a frozen LLM into a recurrent predictor by representing its hidden state as natural-language memory. This state, implemented as a structured system-prompt summary, is updated at each timestep via feedback-driven text rewrites, enabling learning without parameter updates. Under a fixed token budget, LLM-as-RNN corrects errors and retains task-relevant patterns, effectively performing online learning through language. We evaluate the method on three sequential benchmarks in healthcare, meteorology, and finance across Llama, Gemma, and GPT model families. LLM-as-RNN significantly outperforms zero-shot, full-history, and MemPrompt baselines, improving predictive accuracy by 6.5\% on average, while producing interpretable, human-readable learning traces absent in standard context accumulation.
\end{abstract}

\section{Introduction}
Learning from sequential feedback is fundamental to adaptive prediction~\citep{zhang2024large, jiang2024empowering}. Historically, this requirement was met by Recurrent Neural Networks (RNNs) and Long Short-Term Memory (LSTMs), which maintained a compact, evolving hidden state to capture temporal dependencies and adapt to shifting data distributions~\citep{hochreiter1997lstm, cho-etal-2014-learning}. The advent of Transformer architectures revolutionized this landscape by replacing recurrence with large-scale parallel attention mechanisms. In this paradigm, "memory" is no longer a compressed state, but an explicit history of tokens processed via In-Context Learning (ICL)~\citep{brown2020language}. This shift has enabled remarkable advances in reasoning and generation across diverse open and proprietary model families~\citep{wu2025human,du2025rethinking}. 

However, this architectural trade-off introduces a critical limitation in long-horizon settings. During standard inference, Large Language Models (LLMs) operate in a largely stateless way: with frozen parameters, the system lacks a mutable memory to internalize past mistakes ~\citep{shinn2023reflexion, packer2023memgpt}. Instead of updating a belief state, the model relies on an append-only context window, carrying errors forward without correction ~\citep{wang2024wise,muhoberac2025state}. This limitation becomes acute in domains such as longitudinal clinical prediction, weather forecasting, and financial time-series modeling, where task-relevant signals accumulate over time, and the data distribution may drift.

A common solution is to encode the entire past directly in the prompt. One approach, Full History Concatenation (FHC)~\citep{ascoli2025advancing}, appends all raw observations, while methods like MemPrompt~\citep{madaan2022memory} append a step-wise summary. As the sequence grows, concatenation suffers from attention dilution and 'lost-in-the-middle' phenomena~\citep{liu2024lost}, while append-only summaries are prone to error cascading~\citep{pmlr-v235-zhang24ay}. Once a misconception is written into the context, it becomes an immutable ground truth; later evidence often fails to override it, causing errors to persist despite contradictory signals~\citep{turpin2023language}.

\begin{figure*}
    \centering
    \includegraphics[width=\linewidth]{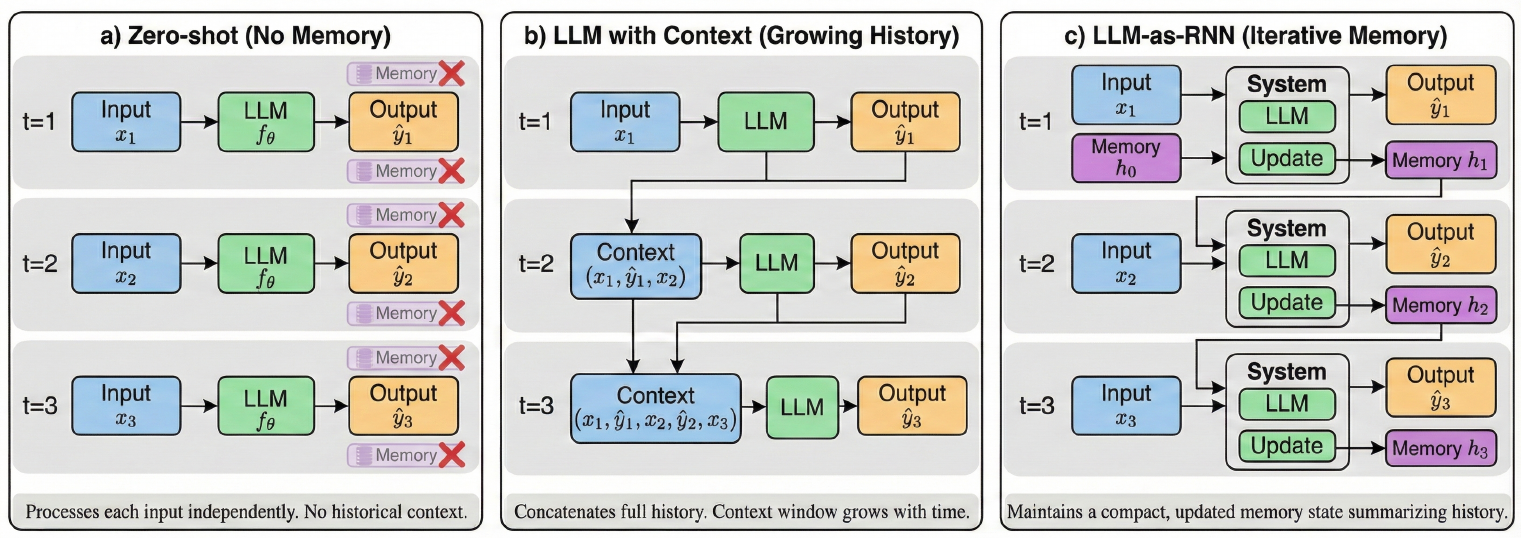}
    \caption{Illustrative comparison. (a) Simple LLM lacks memory. (b) LLM with Context suffers from growing input size. (c) LLM-as-RNN uses an iterative memory state to summarize historical information from evaluating outputs.}
    \vspace{-10pt}
    \label{fig:fig1}
\end{figure*}

We address this gap with LLM-as-RNN, an inference-only framework that reframes sequential prediction as a recurrent process with a natural-language state. Unlike append-only approaches, our method updates a structured memory at each step using feedback, derived from ground-truth labels or an LLM critic, to correct errors and refine strategies under a fixed token budget. We evaluate this approach on three diverse benchmarks: clinical prediction (MIMIC-IV), meteorology (Weather), and financial forecasting (S\&P~500). Across all domains, LLM-as-RNN outperforms zero-shot, full-history concatenation, and MemPrompt baselines, with especially large gains on long sequences. It achieves improvements of 10.8\% on MIMIC-IV, 1.6\% on Weather, and 4.8\% on S\&P 500, while producing interpretable learning traces that make the model’s adaptation process transparent.

This work makes three contributions: (1) We formalize recurrent inference for LLMs, treating textual state as an explicit, mutable memory. This perspective fundamentally distinguishes revisable memory updates from standard unbounded history accumulation.
(2) We introduce LLM-as-RNN, an inference-only framework that enables online adaptation in frozen models. By iteratively rewriting a bounded natural-language state using per-timestep feedback, the model corrects errors without parameter access. (3) We demonstrate across three domains and multiple model families that outcome-driven state updates consistently outperform strong prompt-based baselines. Furthermore, by exposing the adaptation process as human-readable state evolution, our framework facilitates safety audits and builds trust, ensuring that the model's reasoning trajectory is transparent rather than implicit.

\section{Related Work}

\subsection{Recurrent and Memory Models}

Classical sequence modeling uses recurrent neural networks (RNNs), long short-term memory (LSTMs), and gated recurrent units (GRUs) to maintain a vector-valued hidden state that evolves over time~\citep{elman1990finding,hochreiter1997lstm,chung2014empirical,graves2013speech,sutskever2014seq2seq,bahdanau2015neural}. While efficient, these dense vector states often act as an information bottleneck. To address this, memory-augmented architectures, such as Neural Turing Machines and Differentiable Neural Computers, separated the controller from an external differentiable memory bank to support algorithmic reasoning and long-term dependencies~\citep{weston2014memory,graves2014ntm,graves2016dnc,pmlr-v48-santoro16}. 

The Transformer architecture replaced this explicit recurrence with self-attention over a global context~\citep{vaswani2017attention}. While powerful, the quadratic cost of attention has sparked a resurgence of interest in linear-time recurrent architectures. Recent models like RWKV~\citep{peng-etal-2023-rwkv}, Mamba~\citep{gu2024mamba}, and linear attention variants~\citep{katharopoulos2020transformersrnn} effectively reintroduce recurrence into the Transformer backbone, formalizing decoder-only models as multi-state RNNs~\citep{arora2024simple,oren2024transformers}.

However, these approaches typically require training custom architectures from scratch. In the regime of frozen large language models, recurrence is simulated via prompt management. RecurrentGPT~\citep{zhou2023recurrentgpt} and MemGPT~\citep{packer2023memgpt} emulate RNNs by treating the context window as a short-term buffer and offloading history to external storage. LLM-as-RNN builds upon this stateful perspective but distinguishes itself by representing the recurrent state not as a latent vector or a static storage log, but as an evolving, natural-language system prompt. 
\begin{figure*}[t]
    \centering
    \includegraphics[width=\linewidth]{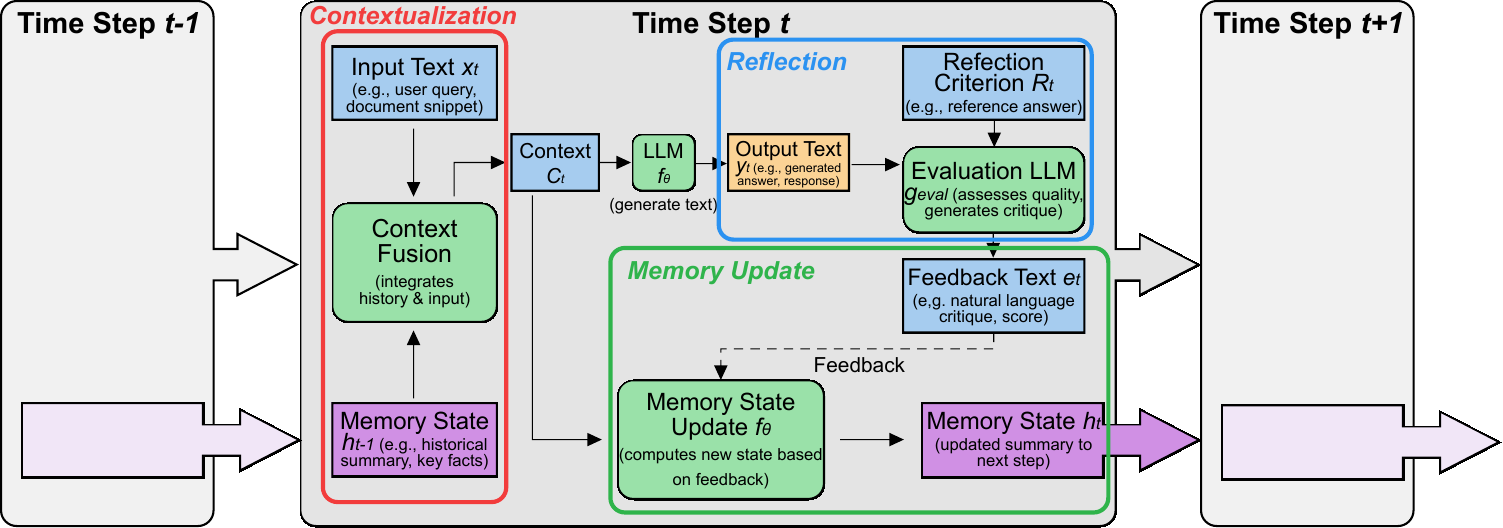}
    \caption{Overview of LLM-as-RNN framework. At each time step, the system fuses the previous memory state with new input to generate a response, evaluates that response to create a feedback signal, and then updates the natural language memory state to guide future interactions.}
    \label{fig:fig2}
\end{figure*}

\subsection{Inference-Time Adaptation in LLMs}

LLMs exhibit strong in-context learning capability, adapting to new tasks from a handful of demonstrations without parameter updates. This behavior has been interpreted as implicit optimization or meta-learning implemented in the forward pass~\citep{brown2020language,akyurek2022iclalgorithm,garg2022can,von2023transformers,dai2023can}. To extend this beyond the context window, retrieval-augmented generation (RAG) and non-parametric systems utilize external memory banks to query relevant history at inference time~\citep{khandelwal2020nearestneighbor,lewis2020retrieval,borgeaud2022improving,izacard2023atlas}. Recent agentic frameworks extend this by storing high-level “events’’ or user profiles to support long-horizon personalization~\citep{park2023generative,zhong2024memorybank,das2024larimar,wang2023augmenting}. However, these approaches are primarily retrieval-based: they select relevant past information but do not necessarily update a belief state to correct errors.

A complementary stream of research treats natural language itself as an optimization variable. Methods like Self-Refine~\citep{madaan2023self}, Reflexion~\citep{shinn2023reflexion}, and ReAct~\citep{yao2022react} introduce iterative feedback loops where the model critiques its own output to improve performance. This concept has been formalized in frameworks like OPRO~\citep{yang2024large} and TextGrad~\citep{yuksekgonul2024textgrad}, which perform "optimization via prompting", effectively backpropagating textual feedback to refine system prompts or solutions.

Unlike RAG, which retrieves static history, and traditional continual learning, which relies on expensive parameter updates~\citep{wu2024continuallearninglargelanguage}, our approach performs adaptation purely at inference time. By treating the system prompt as a recurrent hidden state and updating it through error-driven feedback, we combine recurrent statefulness with feedback-based adaptation in frozen LLMs.

\section{Methods}
\subsection{Preliminaries and Problem Formulation}
We consider sequential generation over a horizon $T$, where the cumulative full history $H_T$ may exceed the LLM's context window $L_{max}$. At each step $t$, the model (parameterized by $\theta$) receives an observation $x_t$, the previous history $H_{t-1} = \{x_{1:t-1}, \hat{y}_{1:t-1}\}$, and generates a response $\hat{y}_t$.
\begin{equation} 
    \hat{y}_t \sim f_\theta(\cdot \mid H_{t-1}, x_t) 
\end{equation}

This formulation suffers from two limitations: (1) Computational: $|H_{t-1}\oplus x_t|$ grows linearly, eventually violating $L_{max}$; and (2) Statelessness: $H_{t-1}$ is an append-only log. Errors in early outputs $\hat{y}_{1:t-1}$ are frozen in the context, permanently biasing future predictions.

To address this, we propose \textbf{LLM-as-RNN} (Figure~\ref{fig:fig2}), which reformulates inference as a recurrent process over a \textit{mutable} textual state $h_t$. Our goal is to maintain a bounded memory $h_{t-1}$, constrained by a fixed token budget $\lambda$ (where $\lambda \ll L_{max}$), that acts as a semantic sufficient statistic for the full history. We seek a memory state that minimizes information loss, ensuring the state-conditioned prediction approximates the full-history:
\begin{equation}
    P_\theta(\hat{y}_t | h_{t-1}, x_t) \approx P_\theta(\hat{y}_t | H_{t-1}, x_t)
\end{equation}

\subsection{Recurrent Inference Mechanism}
The inference process at step $t$ is decomposed into three atomic operations: \textit{Contextualization}, \textit{Reflection}, and \textit{Memory Update}.

\subsubsection{Step 1: Contextualization}
Unlike vector-based RNNs, which fuse inputs via matrix multiplication, LLM-as-RNN performs state mixing directly in the token space. We define a prompt template $\mathcal{P}_{gen}$ that constructs a local context $C_t$ by concatenating ($\oplus$) the system instructions $\mathcal{I}_{sys}$, the prior memory state $h_{t-1}$, and the new observation $x_t$:
\begin{equation}
    C_t = \mathcal{I}_{sys} \oplus h_{t-1} \oplus x_t
\end{equation}
The model then samples a response $\hat{y}_t$ conditioned on this bounded context:
\begin{equation}
    \hat{y}_t \sim f_\theta(\cdot | \mathcal{P}_{gen}(C_t))
\end{equation}

This formulation ensures that the context size remains constant ($O(1)$) with respect to the sequence index $t$. Unlike full-history methods where attention costs grow linearly with time, our input length depends only on the bounded memory size $\lambda$ and current observation $|x_t|$, preventing throughput degradation in long-horizon tasks.

\subsubsection{Step 2: Reflection}
To ensure the memory $h_t$ remains accurate over long horizons, we introduce a feedback mechanism that acts as a "semantic gradient", guiding the evolution of the memory state.
We define a Critic function $g_{eval}$ that evaluates the prediction $\hat{y}_t$ against a reference criterion $\mathcal{R}_t$, producing a natural language feedback signal $e_t$:
\begin{equation}
    e_t = g_{eval}(\hat{y}_t, \mathcal{R}_t)
\end{equation}
We formulate $g_{eval}$ to handle 2 supervision modes:
\begin{itemize}[leftmargin=0pt, label={}]
\item \textbf{Supervised mode:} $\mathcal{R}_t$ contains the ground truth label $y_t$. Here, $g_{eval}$ computes the semantic residual: $e_t \leftarrow \text{``Error: Expected } y_t \text{ but generated } \hat{y}_t \text{.''}$
\item \textbf{Open-Ended mode:} $\mathcal{R}_t$ represents a set of quality heuristics (e.g., relevance, coherence). Here, $g_{eval}$ acts as an \textit{LLM-as-a-Judge}, producing a critique: $e_t \leftarrow \text{``Reasoning flaw: } \hat{y}_t \text{ contradicts prior fact } \dots \text{''}$
\end{itemize}
The signal $e_t$ guides the subsequent memory update, analogous to the backpropagated error term in differentiable memory networks. 

\subsubsection{Step 3: Memory Update}
The final step is the memory update, where we transition from $h_{t-1} \to h_t$ to incorporate new information while satisfying the token budget $\lambda$. This operation is modeled not just as summarization, but as a feedback-guided rewrite.

Using a specific prompt template $\mathcal{P}_{mem}$, the model generates the new state conditioned on the previous state, current events, and the feedback signal $s_t$:
\begin{equation}
h_t \sim f_\theta(\cdot \mid \mathcal{P}_{mem}(h_{t-1}, x_t, \hat{y}_t, e_t))
\end{equation}
The prompt $\mathcal{P}_{mem}$ explicitly instructs the model to:
\begin{enumerate}
\item Compress $x_t$ and $\hat{y}_t$ into the summary.
\item Apply the Critique: Use $e_t$ to identify and rewrite incorrect beliefs in $h_{t-1}$ rather than simply appending new tokens.
\end{enumerate}
This formulation ensures the memory is self-healing: the state evolves to correct misconceptions based on the "semantic gradient" $e_t$.

\subsection{Algorithm}
The complete inference procedure, integrating the semantic gradient loop and constraint enforcement, is detailed in Algorithm 1.

\begin{algorithm}[h]
\caption{LLM-as-RNN Inference Process}
\label{alg:llm_rnn}
\begin{algorithmic}[1]
\REQUIRE Sequence $\{x_t\}_{t=1}^T$, Frozen LLM $\theta$, Evaluation Function $g_{eval}$, Max Memory $\lambda$
\STATE $h_0 \leftarrow \varnothing$
\FOR{$t = 1$ to $T$}
    \STATE \textit{// Phase 1: Contextualization}
    \STATE $C_t \leftarrow \mathcal{I}_{sys} \oplus h_{t-1} \oplus x_t$
    \STATE $\hat{y}_t \sim f_\theta(\cdot | \mathcal{P}_{gen}(C_t))$ \COMMENT{Generate Prediction}
    
    \STATE \textit{// Phase 2: Reflection}
    \STATE Retrieve reference/criteria $\mathcal{R}_t$
    \STATE $e_t \leftarrow g_{eval}(\hat{y}_t, \mathcal{R}_t)$ \COMMENT{Semantic Evaluation}
    
    \STATE \textit{// Phase 3: Memory Update}
    \STATE Update $h_t \leftarrow f_\theta(\cdot | \mathcal{P}_{mem}(h_{t-1}, x_t, \hat{y}_t, e_t))$
    % \STATE $h_t \leftarrow \text{Generate}(h_{t-1}; \phi)$
    \IF{$|h_t| > \lambda$}
        \STATE $h_t \leftarrow \text{Compress}(h_t, \lambda)$
    \ENDIF
\ENDFOR
\RETURN $\{\hat{y}_t\}_{t=1}^T$
\end{algorithmic}
\end{algorithm}

\section{Experiments}

\subsection{Datasets}
\label{sec:datasets}

We evaluate LLM-as-RNN on three sequential benchmarks spanning healthcare, meteorology, and finance: the MIMIC-IV dataset, the Weather dataset, and the S\&P 500 with Financial News Headlines dataset. 
Following a unified protocol, we structure the weather and finance benchmarks as continuous temporal streams rather than independent samples.
Additional dataset statistics and preprocessing details are provided in Appendix~\ref{sec:dataset_details}.

\paragraph{MIMIC-IV.}
The MIMIC-IV dataset is a deidentified EHR dataset containing ICU admissions with structured clinical variables (e.g., diagnoses, procedures, labs, treatments)~\citep{johnson2023mimic}. We construct longitudinal patient trajectories consisting of sequential clinical notes and lab events (Appendix~\ref{sec:mimiciv_filtering}). The task is to predict the primary diagnosis for the next hospital visit given the history of prior admissions.

\paragraph{Weather.}
The Weather dataset is a meteorological time series containing mixed modalities, including textual descriptors and continuous measurements (e.g., temperature, humidity, wind, pressure)~\citep{kaggle_weather_dataset_muthuj7}. We employ a \textit{sequential sliding-window protocol}: at each timestep $t$, the observation $x_t$ is restricted to a fixed 5-day window. However, the model predicts the condition for day $t$ by conditioning on both this local window \textit{and} the recurrent memory $h_{t-1}$. 

\paragraph{S\&P 500.}
The S\&P 500 dataset aligns daily market closing prices with financial news headlines~\citep{kaggle_sp500_news_dyutidasmahaptra}. Under the same protocol, the model receives a 5-day lookback of price and news as input $x_t$. The task is to forecast the closing price for day $t$, synthesizing quantitative signals with qualitative sentiment accumulated over the entire sequence.

\subsection{Baselines}
We compare LLM-as-RNN against three baselines. All methods share the same input/output formatting and evaluation protocol; they differ only in how they encode history.

\paragraph{Zero-shot.}
This baseline (Figure~\ref{fig:fig1}a) predicts the target at time $t$ using only the most recent observation (e.g., the last visit/day) without any additional history. It serves as a lower bound that tests the LLM’s raw single-step predictive ability.

\paragraph{Full History Concatenation (FHC).}
FHC~\citep{ascoli2025advancing} (Figure~\ref{fig:fig1}b) conditions the LLM on the entire available history by directly concatenating all past observations into the prompt at each timestep:
\begin{equation}
C_t = x_1 \oplus x_2 \oplus \cdots \oplus x_{t-1} \oplus x_t.
\end{equation}
FHC is the most straightforward strategy for leveraging temporal context and is commonly used as a long-context baseline, but its input length grows with $t$ and can exceed the model’s context window, necessitating truncation and potentially degrading performance.

\paragraph{MemPrompt.}
MemPrompt~\citep{madaan2022memory} summarizes each past observation into a short textual memory unit and concatenates the accumulated summaries as a compact proxy for the full history:
\begin{equation}
\begin{aligned}
m_i &= \text{Summarize}(x_i), \\
C_t &= m_1 \oplus m_2 \oplus \dots \oplus m_{t-1} \oplus x_t.
\end{aligned}
\end{equation}
Unlike FHC, MemPrompt bounds history growth via per-step compression. However, the memory is append-only, previously written summaries are not revised in light of new evidence or prediction errors, so misconceptions can persist over time.

\begin{table*}[t]
  \centering
  \caption{\textbf{Overall performance.} \textbf{Green} = best backbone within each method; \textbf{Yellow} = best overall in the full table.}
  \label{tab:main-results}
  \setlength{\tabcolsep}{7.6pt}
  \renewcommand{\arraystretch}{0.4} % slightly less cramped than 0.5
  \begin{tabular}{l l cc c cc}
    \toprule
    \rowcolor{headerblue}
    \multirow{2}{*}{\textbf{Method}} &
    \multirow{2}{*}{\textbf{LLM backbones}} &
    \multicolumn{2}{c}{\textbf{MIMIC-IV}} &
    \textbf{Weather} &
    \multicolumn{2}{c}{\textbf{S\&P 500}} \\
    \cmidrule(lr){3-4}\cmidrule(lr){5-5}\cmidrule(lr){6-7}
    & & Acc@1($\uparrow$) & Acc@5($\uparrow$) & Align($\uparrow$) & MAE($\downarrow$) & MSE($\downarrow$) \\
    \midrule

    % ---------------- Zero-shot ----------------
    \multirow{15}{*}{\textbf{Zero-shot}} & Llama-3.2-1B   & 0.1538 & 0.3846 & 0.6783 & 1.598 & 6.273 \\
    & Llama-3.2-3B   & 0.1958 & 0.4545 & 0.6923 & 1.585 & 6.080 \\
    & Llama-3.1-8B   & 0.2797 & 0.5594 & 0.7063 & 1.499 & 5.343 \\
    & Llama-3.1-70B  & 0.3287 & 0.6503 & 0.7203 & 1.493 & 5.232 \\
    & Gemma-3-1B   & 0.1608 & 0.3986 & 0.6853 & 1.612 & 6.347 \\
    & Gemma-3-4B   & 0.2238 & 0.5175 & 0.7063 & 1.548 & 5.982 \\
    & Gemma-3-12B  & 0.2937 & 0.6014 & 0.7273 & 1.438 & 5.214 \\
    & Gemma-3-27B  & 0.3427 & 0.6713 & \best{0.7552} & 1.402 & 4.972 \\
    & GPT-oss-20B  & 0.3357 & 0.6643 & 0.7413 & \best{1.305} & \best{4.590} \\
    & GPT-oss-120B & \best{0.3636} & \best{0.7063} & 0.7483 & 1.312 & 4.612 \\
    \midrule

    % ---------------- FHC ----------------
    \multirow{15}{*}{\textbf{FHC}} & Llama-3.2-1B   & 0.2517 & 0.6434 & 0.7133 & 1.609 & 5.738 \\
    & Llama-3.2-3B   & 0.3357 & 0.7552 & 0.7133 & 1.578 & 5.503 \\
    & Llama-3.1-8B   & 0.3497 & 0.8392 & 0.7203 & 1.440 & 5.317 \\
    & Llama-3.1-70B  & 0.4126 & 0.8601 & 0.7273 & 1.464 & 5.211 \\
    & Gemma-3-1B   & 0.2587 & 0.6503 & 0.7133 & 1.603 & 6.042 \\
    & Gemma-3-4B   & 0.3566 & 0.8112 & 0.7273 & 1.511 & 5.541 \\
    & Gemma-3-12B  & 0.4056 & 0.8601 & 0.7483 & 1.412 & 5.067 \\
    & Gemma-3-27B  & \best{0.4476} & \best{0.9021} & 0.7832 & 1.386 & 4.881 \\
    & GPT-oss-20B  & 0.4126 & 0.8671 & 0.7902 & \best{1.280} & \best{4.420} \\
    & GPT-oss-120B & 0.4406 & 0.8881 & \best{0.7972} & 1.290 & 4.440 \\
    \midrule

    % ---------------- MemPrompt ----------------
    \multirow{15}{*}{\textbf{MemPrompt}} & Llama-3.2-1B   & 0.3147 & 0.7203 & 0.7203 & 1.478 & 5.879 \\
    & Llama-3.2-3B   & 0.4126 & 0.7902 & 0.7273 & 1.426 & 5.148 \\
    & Llama-3.1-8B   & 0.4336 & 0.8462 & 0.7343 & 1.309 & 4.768 \\
    & Llama-3.1-70B  & 0.4615 & \best{0.9231} & 0.7413 & 1.292 & 4.417 \\
    & Gemma-3-1B   & 0.3217 & 0.7273 & 0.7273 & 1.491 & 6.118 \\
    & Gemma-3-4B   & 0.4196 & 0.8392 & 0.7483 & 1.392 & 5.012 \\
    & Gemma-3-12B  & 0.4755 & 0.8951 & 0.7832 & 1.298 & 4.716 \\
    & Gemma-3-27B  & 0.4965 & 0.9091 & 0.8182 & 1.264 & \best{4.090} \\
    & GPT-oss-20B  & 0.4895 & 0.9021 & 0.8112 & 1.230 & 4.120 \\
    & GPT-oss-120B & \best{0.5175} & 0.9161 & \overallbest{0.8322} & \best{1.198} & 4.162 \\
    \midrule

    % ---------------- LLM-as-RNN ----------------
    \multirow{15}{*}{\textbf{\shortstack{LLM-as-RNN\\(Ours)}}} & Llama-3.2-1B   & 0.3427 & 0.7552 & 0.7343 & 1.531 & 5.297 \\
    & Llama-3.2-3B   & 0.4545 & 0.7972 & 0.7413 & 1.347 & 5.025 \\
    & Llama-3.1-8B   & 0.5524 & 0.8531 & 0.7483 & 1.287 & 4.517 \\
    & Llama-3.1-70B  & 0.5804 & \overallbest{0.9510} & 0.7622 & 1.211 & 4.313 \\
    & Gemma-3-1B   & 0.3497 & 0.7622 & 0.7413 & 1.533 & 5.541 \\
    & Gemma-3-4B   & 0.4685 & 0.8601 & 0.7622 & 1.321 & 4.754 \\
    & Gemma-3-12B  & 0.5594 & 0.9161 & \best{0.8252} & 1.237 & 4.406 \\
    & Gemma-3-27B  & \overallbest{0.6434} & 0.9301 & 0.8182 & 1.186 & 4.072 \\
    & GPT-oss-20B  & 0.5874 & 0.9231 & 0.8112 & \overallbest{1.105} & 3.900 \\
    & GPT-oss-120B & 0.6294 & 0.9441 & 0.8182 & 1.112 & \overallbest{3.821} \\
    \bottomrule
  \end{tabular}
\end{table*}

\subsection{LLM Backbones}
We evaluate LLM-as-RNN with multiple backbone LLMs to assess how backbone capability affects sequential prediction performance (Llama~\citep{grattafiori2024llama}, Gemma~\citep{team2025gemma}, and GPT~\citep{agarwal2025gpt} families). To ensure fair comparisons across backbones, we use \textit{temperature} $=0.7$, \textit{top-p} $=0.9$, and \textit{max\_tokens} $=4096$ for all LLM calls, and keep these settings unchanged across timesteps in the sequential stream. All LLM-based evaluations are computed using the same judge model, \textit{Claude Sonnet 4.5}~\citep{anthropic2025claudeSonnet45}.

\subsection{Metrics}
We employ task-specific metrics: 
\begin{itemize}[leftmargin=*] 
    \item \textbf{MIMIC-IV (Semantic Accuracy):} We report LLM-Judged Accuracies (Acc@1 and Acc@5). The LLM-Judge determines if the generated diagnosis is semantically equivalent to the ground truth, avoiding the pitfalls of string matching. 
    \item \textbf{Weather (Alignment):} The LLM judge evaluates whether the generated summary factually contradicts the ground truth on key variables, producing a binary success score. 
    \item \textbf{S\&P 500 (Forecasting Error):} We measure the deviation between predicted and actual closing prices using Mean Absolute Error (MAE) and Mean Squared Error (MSE). 
\end{itemize}

\begin{figure*}
    \centering
    \includegraphics[width=\textwidth]{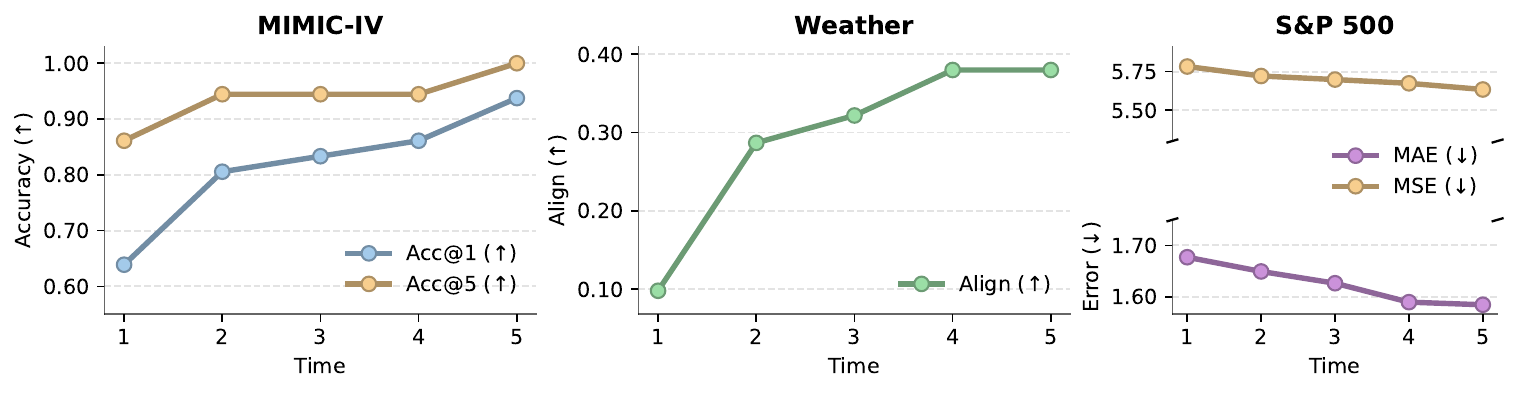}
    \vspace{-18pt}
    \caption{Temporal dynamics across iterative timesteps ($t{=}1\ldots5$) for three datasets. As feedback-driven memory updates accumulate, the performance increase.}
    \vspace{-10pt}
    \label{fig:temporal_dynamics}
\end{figure*}

% Result-1: Overall results analysis (comparison with baselines and different LLM backbones)

% Result-2: With the increase of past visits, the performance increases

% Result-3: When replace the ground truth with LLM-as-a-judge, the performance still stay strong. 

% Result-4: Ablation study. 

\section{Results}
\label{sec:results}

Our evaluation is guided by 4 research questions:

\begin{itemize}[leftmargin=*]
    \item \textbf{RQ1 (Efficacy):} Does the LLM-as-RNN framework outperform other strategies, and how does different LLM backbones behave?
    \item \textbf{RQ2 (Temporal Dynamics):} Does the model learn and reduce error over long sequences?
    \item \textbf{RQ3 (Autonomy):} Can the framework operate effectively using intrinsic self-correction (LLM-as-a-Judge) in the absence of ground truth labels?
    \item \textbf{RQ4 (Mechanism):} What are the contributions of the memory to the overall performance?
\end{itemize}

\subsection{Overall Performance Analysis (RQ1)}

% ================================
We analyze whether outcome-driven memory updates can effectively overcome the shortcomings of static context accumulation. As shown in Table~\ref{tab:main-results}, LLM-as-RNN achieves the strongest overall performance across most backbones and tasks, supporting our core hypothesis: achieving robustness over long horizons requires a mutable belief state rather than a monotonically growing context buffer.

The advantage is largest in settings where the ground truth changes over time or conflicts with earlier evidence. On MIMIC-IV, where a patient’s condition can change abruptly, LLM-as-RNN improves Acc@1 by 12.6\% (absolute) compared to the best MemPrompt variant (0.6434 vs. 0.5175). Similarly, on S\&P 500, where shifts in market regimes invalidate prior “trends,” our method lowers MSE by roughly 6.6\% (3.821 vs. 4.090). These results suggest that standard append-only methods (FHC and MemPrompt) are prone to error persistence: once an incorrect diagnosis or outdated sentiment is written into the context, it continues to bias future predictions. In contrast, LLM-as-RNN can actively “forget” or revise these obsolete patterns through its update mechanism.

A direct comparison between LLM-as-RNN and FHC underscores the inefficiency of raw context. Although FHC receives a complete, lossless history, it consistently underperforms our compressed-state approach (e.g., for Llama-3-70B, FHC Acc@1 is 0.4126 vs. 0.5804 for LLM-as-RNN). This indicates that the bottleneck in sequential prediction is not how much information is available, but how well that information is curated. FHC is vulnerable to attention dilution and noise, whereas LLM-as-RNN acts as an information filter that preserves only the signals relevant for the next prediction.

Although performance generally improves with model size, LLM-as-RNN disproportionately benefits smaller backbones. On the clinical prediction task, the Llama-3.2-3B model with LLM-as-RNN (Acc@1: 0.4545) surpasses the significantly larger Llama-3.1-70B with FHC (Acc@1: 0.4126). This suggests that the recurrent update loop acts as a strong inductive bias for sequential reasoning, enabling smaller models to approximate the long-horizon tracking capabilities typically associated with many more parameters.

On the Weather benchmark, while LLM-as-RNN outperforms baselines for most backbones (7/10), the gains are smaller, and MemPrompt achieves the single highest alignment score (0.8322 with GPT-oss-120B). We hypothesize that meteorological data, which is physically continuous and less semantic, derives limited benefit from “verbal correction” compared to semantic tasks such as diagnosis. In these high-entropy physical processes, an append-only memory scheme like MemPrompt may already be adequate, since weather trajectories seldom exhibit the sort of logical inconsistencies or conceptual errors that our textual update mechanism is designed to repair.

\subsection{Temporal Dynamics and Learning (RQ2)}
A key hypothesis of LLM-as-RNN is that the framework does not merely condition on history but \textit{learns} from it through recurrence. To validate this, we analyze the model's performance as a function of the sequence length $t$.

\paragraph{Performance Gains Over Time.}
Figure~\ref{fig:temporal_dynamics} shows a clear improvement trend as LLM-as-RNN updates its recurrent memory state across timesteps. Importantly, at each timestep $t$ we evaluate the model’s prediction $\hat{y}_t$ against the corresponding ground-truth target $y_t$ for that timestep. The reported curves reflect how performance evolves from the first prediction through later steps using feedback from previous steps.

On MIMIC-IV, both Acc@1 and Acc@5 increase substantially from early to late time steps, with the largest jump occurring after the first update and continued gains thereafter. On Weather, alignment rises rapidly in the first few steps and then plateaus, suggesting the state quickly captures the key short-horizon signals. On S\&P 500, MAE and MSE decrease steadily but more modestly, indicating incremental calibration of numerical forecasts. Overall, these curves support the core hypothesis that feedback-driven state rewrites enable online improvement under a fixed budget, rather than merely accumulating history.

\paragraph{Recovery from Error.}
We quantify recovery from error as an incorrect to correct transition between consecutive visits (Appendix Figure~\ref{fig:transition}). Conditioned on being incorrect at time $t$, LLM-as-RNN corrects its prediction at $t{+}1$ in 54.8\% of cases after incorporating the feedback signal $e_t$, while 45.2\% of errors persist. Overall, these transition dynamics support a feedback-driven ``self-healing'' behavior enabled by memory updates, which is harder to obtain in static baselines that lack an explicit recurrent state update.

\subsection{Robustness to Feedback Source (RQ3)}
The standard configuration of LLM-as-RNN relies on ground-truth supervision to generate the feedback signal $e_t$. However, in many real-world deployment scenarios, immediate ground truth is unavailable. We evaluate the more common ``Open-Ended'' mode where the feedback $e_t$ is generated by an LLM-Critic. This critic is grounded in a specific set of domain guidelines.

\begin{table}[t]
  \centering
  \small
  \caption{Performance comparison between supervised feedback (ground-truth) and self-supervised feedback (LLM-as-a-Judge). Results use Llama-3.1-8B.}
  \label{tab:llm-as-a-judge}
  \setlength{\tabcolsep}{4.5pt}
  \renewcommand{\arraystretch}{1}
  \begin{tabular}{l l ccc}
    \toprule
    \textbf{Dataset} & \textbf{Metric} & \textbf{Ground Truth} & \textbf{LLM-Judge} \\
    \midrule
    \multirow{2}{*}{MIMIC-IV} 
      & Acc@1 ($\uparrow$) & 0.5524 & 0.3077 \\
      & Acc@5 ($\uparrow$) & 0.8531 & 0.8042 \\
    \midrule
    Weather 
      & Align ($\uparrow$) & 0.7483 & 0.7203 \\
    \midrule
    \multirow{2}{*}{S\&P 500} 
      & MAE ($\downarrow$) & 1.287 & 1.545 \\
      & MSE ($\downarrow$) & 4.517 & 5.533 \\
    \bottomrule
  \end{tabular}
\end{table}

Table~\ref{tab:llm-as-a-judge} shows the performance gap when replacing the ground truth $g_{eval}(\hat{y}_t, y_t)$ with a neural critic $g_{eval}(\hat{y}_t, \mathcal{R}_t)$. Overall, switching to open-ended, self-supervised feedback degrades performance, but still achieving acceptable results. On MIMIC-IV, Acc@1 drops from 0.5524 to 0.3077, while Acc@5 remains comparatively robust, suggesting that the LLM-Judge can preserve a high-quality candidate set but provides a weaker fine-grained ranking signal. On the Weather dataset, the Alignment Rate decreases marginally from 0.7483 to 0.7203, retaining 96.3\% of the fully supervised performance. On the S\&P 500 forecasting task, the degradation is larger: MAE increases from 1.287 to 1.545 and MSE increases from 4.517 to 5.533. Nevertheless, the model with LLM judges significantly outperforms the zero-shot baseline, validating that the iterative reflection process itself, even without perfect labels, induces a form of self-consistency that stabilizes long-horizon generation.

\subsection{Ablation Studies (RQ4)}
To isolate the effect of the context window budget available to the recurrent state in LLM-as-RNN, we conducted a controlled ablation on the S\&P 500 dataset. Specifically, we varied the maximum context window length $\lambda$ for each LLM call across 512, 1024, 2048, 4096, and 8192, and report forecasting errors in Table~\ref{tab:ablation_memory}.

As $\lambda$ increases, both MAE and MSE consistently decrease, indicating that allocating a larger context window to the recurrent state enables the model to retain more relevant historical signals and reduce prediction error. The gains are most pronounced when scaling from $\lambda=512$ to $\lambda=4096$ (MAE: $1.564 \rightarrow 1.287$, MSE: $5.462 \rightarrow 4.517$). Further increasing the budget to $\lambda=8192$ yields only marginal additional improvement (MAE: $1.287 \rightarrow 1.248$, MSE: $4.517 \rightarrow 4.366$), suggesting diminishing returns at larger context windows. Overall, these results support the importance of sufficient context window budget for trajectory modeling while indicating that performance saturates once $\lambda$ is large enough to preserve the time-series signals.

\begin{table}[t]
  \centering
  \small
  \caption{Effect of context window token budget $\lambda$ on S\&P 500 dataset using Llama-3.1-8B. Performance improves with increased budget up to a threshold, beyond which additional capacity yields diminishing returns.}
  \label{tab:ablation_memory}
  \setlength{\tabcolsep}{13.5pt}
  \renewcommand{\arraystretch}{1.1}
  \begin{tabular}{c cc}
    \toprule
    \textbf{Context window} $\boldsymbol{\lambda}$ & \textbf{MAE} ($\downarrow$) & \textbf{MSE} ($\downarrow$) \\
    \midrule
    512  & 1.564 & 5.462 \\
    1024  & 1.499 & 5.292 \\
    2048  & 1.443 & 5.167 \\
    4096 & 1.287 & 4.517 \\
    8192 & 1.248 & 4.366 \\
    \bottomrule
  \end{tabular}
\end{table}

\section{Conclusion}
We presented LLM-as-RNN, an inference-time framework that makes a frozen LLM behave like a recurrent predictor by maintaining a natural-language memory state and rewriting it at each timestep using feedback on prior predictions, rather than accumulating an immutable context history. This revisable, structured state enables online error correction under a fixed token budget while producing transparent, human-readable learning traces. Across time sequential benchmarks in healthcare, meteorology, and finance, our results indicate that feedback-driven state rewriting offers a simple, model-agnostic route to long-horizon adaptation without parameter updates.

\clearpage

\section{Limitation}
LLM-as-RNN increases inference cost because each timestep typically requires multiple model calls (prediction, reflection, and memory update), which can be prohibitive for latency-sensitive deployments compared to single-pass inference. The quality of memory updates is bounded by the backbone model’s ability to diagnose failures, translate errors into actionable guidance, and compress information without losing critical signals; smaller backbones can be more prone to unstable or lossy updates. The framework also depends on the availability and reliability of feedback: ground-truth labels may be delayed or unavailable, and LLM-based critics can introduce noise or inconsistency that leads to memory drift or self-reinforcing mistakes. Finally, while the core algorithm is domain-agnostic, strong performance often requires careful prompt/schema design and selecting an appropriate memory budget, and fully automating prompt design and ensuring robust long-horizon behavior remain open challenges.

\section{Potential risks}
Because LLM-as-RNN accumulates state over time, incorrect predictions or memory updates can compound across timesteps, and the interpretability of the memory may create an illusion of reliability that increases automation bias; in clinical decision support, the system should be used strictly as an assistive tool with clinician oversight and rigorous evaluation across diverse subpopulations, and outputs should not be treated as medical advice. Similar concerns apply to financial forecasting: markets are influenced by exogenous factors and regime shifts, and sequential “learning from feedback” can overfit to noise or recent trends, so the framework should not serve as the sole basis for investment decisions. The memory mechanism also raises privacy and security concerns because the state may inadvertently retain sensitive or re-identifying details if not carefully controlled, and untrusted inputs or adversarial feedback could corrupt the memory and steer future predictions; mitigations include data-minimization and redaction for stored state, access controls, separating trusted feedback channels from user content, enforcing structured update schemas, monitoring for anomalous memory changes, and providing reset/rollback mechanisms.

\bibliography{custom}

\clearpage

\appendix

\section{Dataset Details}
\label{sec:dataset_details}

\paragraph{MIMIC-IV (clinical EHR).}
MIMIC-IV~\footnote{https://physionet.org/content/mimiciv/3.1/}~\citep{johnson2023mimic} is a large, deidentified electronic health record (EHR) database of patients treated at the Beth Israel Deaconess Medical Center, covering both intensive care unit (ICU) admissions and emergency department (ED) visits. It contains structured clinical information such as demographics, diagnoses, procedures, laboratory measurements, and treatment/medication-related variables. In our experiments, we represent each patient as a chronologically ordered sequence of visits/admissions, and we apply deterministic visit filtering (Appendix~\ref{sec:mimiciv_filtering}) to reduce clinically heterogeneous timelines.

\paragraph{Weather (meteorological time series).}
The Weather dataset~\footnote{https://www.kaggle.com/datasets/muthuj7/weather-dataset} is a multivariate time series with 96{,}453 timestamped observations and 12 columns, mixing categorical descriptors and continuous meteorological variables. Typical fields include a timestamp (Formatted Date), textual descriptors (e.g., Daily Summary), precipitation type, and numeric measurements such as temperature and apparent temperature ($^\circ$C), humidity, wind speed, wind bearing, visibility, and pressure.

\paragraph{S\&P 500 with Financial News Headlines (financial time series).}
S\&P 500 dataset~\footnote{https://www.kaggle.com/datasets/dyutidasmahaptra/s-and-p-500-with-financial-news-headlines-20082024} dataset couples a historical S\&P 500 market time series closing prices with one or more daily financial news headlines. We align market records and headlines by trading date. This benchmark emphasizes non-stationarity and temporal dependence typical of financial markets, while also testing whether textual news can help guide sequential prediction and memory updates.

\begin{table*}[t]
  \centering
  \small
  \caption{Dataset summary (appendix). Fill in cohort-specific counts after preprocessing where applicable.}
  \label{tab:dataset_summary}
  \setlength{\tabcolsep}{5pt}
  \renewcommand{\arraystretch}{1.2}
  \begin{tabular}{l p{1.6cm} p{1.3cm} p{4cm} p{2.6cm} p{2.8cm}}
    \toprule
    \textbf{Dataset} & \textbf{Domain} & \textbf{Timestep} &
    \textbf{Inputs (modalities)} &
    \textbf{Prediction target} &
    \textbf{Metric(s)} \\
    \midrule
    MIMIC-IV &
    Healthcare &
    Visit &
    Structured EHR notes (labs, diagnoses, treatments) &
    Clinical diagnosis &
    Acc@1, Acc@5 ($\uparrow$) \\
    \midrule
    Weather &
    Meteorology &
    Hourly &
    Numeric variables and categorical / text descriptors &
    Weather summary&
    Alignment Rate ($\uparrow$) \\
    \midrule
    S\&P 500 &
    Finance &
    Daily&
    News headlines text &
    Closing prices &
    MAE, MSE ($\downarrow$) \\
    \bottomrule
  \end{tabular}
\end{table*}

% Optional: If you want a second table with dataset sizes, keep it separate to avoid width issues.
% Fill in the placeholders once you finalize preprocessing.
\begin{table}[t]
  \centering
  \small
  \caption{Dataset statistics.}
  \label{tab:dataset_sizes}
  \setlength{\tabcolsep}{10pt}
  \renewcommand{\arraystretch}{1.2}
  \begin{tabular}{l c}
    \toprule
    \textbf{Dataset} & \textbf{Size} \\
    \midrule
    MIMIC-IV & 7,128 patients,\ 37,536 visits \\
    Weather  & 4,019 days,\ 96,453 observations \\
    S\&P 500 & 3,507 trading days,\ 19,127 headlines \\
    \bottomrule
  \end{tabular}
\end{table}

\section{Visit Filtering for MIMIC-IV}
\label{sec:mimiciv_filtering}

Patient timelines in MIMIC-IV may contain admissions that are clinically heterogeneous across time (e.g., unrelated comorbid events). We apply a deterministic, lexicon-driven filtering procedure that retains, for each patient, a temporally contiguous subsequence of visits whose inferred coarse topics are mutually consistent. The procedure does not learn from labels or train a model; it only prunes visits while preserving all original structured fields of the retained records.

\subsection{Inputs, Outputs, and Cohort Preselection}
\paragraph{Inputs.}
The filtering consumes the full parsed MIMIC-IV dataset in JSON format, where each patient record contains a chronologically ordered list of visit records.

\paragraph{Cohort restriction.}
Only patients whose \texttt{valid\_visits} fall within a fixed range are considered. In the reported configuration, we restrict to patients with 5--20 valid visits (inclusive).

\paragraph{Outputs.}
The procedure produces a filtered JSON dataset in which each retained patient record preserves all original fields but replaces the visit list with the selected subsequence.

\subsection{Visit Text Construction}
For each visit, we construct a single lowercased text string by concatenating multiple free-text sources:
\begin{itemize}[leftmargin=*]
    \item \textbf{Clinical sections:} all section values; if a section value is a list, all elements are included.
    \item \textbf{Notes:} if a structured notes field is present, the full note text is used when available.
    \item \textbf{Additional fields:} like \texttt{chief\_complaint}, \texttt{allergies} and \texttt{service}.
\end{itemize}
This aggregated text is used only for topic matching, it does not alter the stored visit content.

We define a small set of coarse medical topics, each represented by a set of keywords. Topic evidence is computed by substring matches in the aggregated visit text. The full lexicon is shown in Table~\ref{tab:topic_lexicon}.

\begin{table}[t]
\centering
\small
\renewcommand{\arraystretch}{1.5}
\caption{Coarse topic lexicon used for visit topic assignment. Abbreviations such as CHF, COPD, HbA1c, CKD, and AKI are matched as substrings.}
\begin{tabular}{l m{0.63\linewidth}}
\hline
\textbf{Topic} & \textbf{Keywords} \\
\hline
Cardiovascular &
heart, cardiac, cardiovascular, hypertension, chf, myocardial, coronary, artery, atrial, ventricular, angina, infarction, fibrillation, blood pressure \\
Respiratory &
lung, pulmonary, respiratory, pneumonia, copd, asthma, bronchitis, dyspnea, breathing, oxygen \\
Diabetes &
diabetes, diabetic, glucose, insulin, hyperglycemia, hypoglycemia, hba1c, blood sugar \\
Renal &
kidney, renal, nephro, dialysis, creatinine, ckd, aki, urinary, urine \\
Neurological &
neuro, brain, stroke, seizure, dementia, alzheimer, parkinson, headache, migraine \\
Gastrointestinal &
gastro, intestinal, liver, stomach, bowel, gi, hepatic, cirrhosis, abdominal \\
Oncology &
cancer, tumor, malignancy, metastasis, oncology, chemotherapy, radiation, neoplasm \\
Infectious &
infection, sepsis, bacterial, viral, antibiotic, fever, inflammatory \\
\hline
\end{tabular}
\label{tab:topic_lexicon}
\end{table}

\subsection{Per-Visit Topic Assignment}
Let $v$ denote a visit and $t$ a topic with keyword set $K_t$. We compute a topic score as the number of keywords that appear in the visit text:
\begin{equation}
    \small
s_t(v) \;=\; \sum_{k \in K_t} \mathbf{1}\big[k \text{ in the aggregated text of } v\big]
\end{equation}
Topics with $s_t(v)=0$ are ignored. The visit is assigned up to the top three topics by $s_t(v)$ (ties broken by the sorting order):
\begin{equation}
\mathrm{Topics}(v) \;=\; \mathrm{Top}\text{-}3\{t : s_t(v)>0\}.
\end{equation}
If no keywords match, then $\mathrm{Topics}(v)=\emptyset$.

For two visits $v_i$ and $v_j$, we compute Jaccard similarity over topic sets:
\begin{equation}
\small
\mathrm{Sim}(v_i, v_j) \;=\;
\begin{cases}
\dfrac{|\mathrm{Topics}(v_i)\cap \mathrm{Topics}(v_j)|}{|\mathrm{Topics}(v_i)\cup \mathrm{Topics}(v_j)|}
\end{cases}
\end{equation}
This definition ensures that visits with no matched topics do not spuriously increase coherence.

\subsection{Consecutive Group Discovery}
For each patient with visits $(v_1,\dots,v_n)$ in chronological order, we partition the sequence into \emph{consecutive} groups using a single left-to-right pass.

We maintain a current group $G$ (initialized with the first visit). For each subsequent visit $v_i$, we compute its average similarity to the visits already in the current group:
\begin{equation}
\overline{\mathrm{Sim}}(v_i, G) \;=\; \frac{1}{|G|}\sum_{v_j \in G} \mathrm{Sim}(v_i, v_j).
\end{equation}
If $\overline{\mathrm{Sim}}(v_i, G) \ge \tau$, we append $v_i$ to $G$; otherwise we close $G$ and start a new group at $v_i$. Only groups of length at least $m$ are kept as candidate coherent groups. If no candidate group exists (i.e., no consecutive segment reaches length $m$), we fall back to treating the entire visit sequence as a single group.

Given the candidate coherent groups for a patient, we select which visits to keep by retaining the largest group. The resulting filtered visit list is the chronologically ordered subsequence corresponding to that group.

Finally, we enforce a minimum number of retained visits: if a patient has fewer than $r$ visits after filtering, the patient is removed from the filtered dataset. In our experiments, we use $\tau=0.6$, $m=2$ and $r=3$.

For all retained patients, all original patient-level fields are preserved unchanged; only the visit list is replaced by the selected subsequence.

\begin{figure*}[t]
    \centering
    \includegraphics[width=\textwidth]{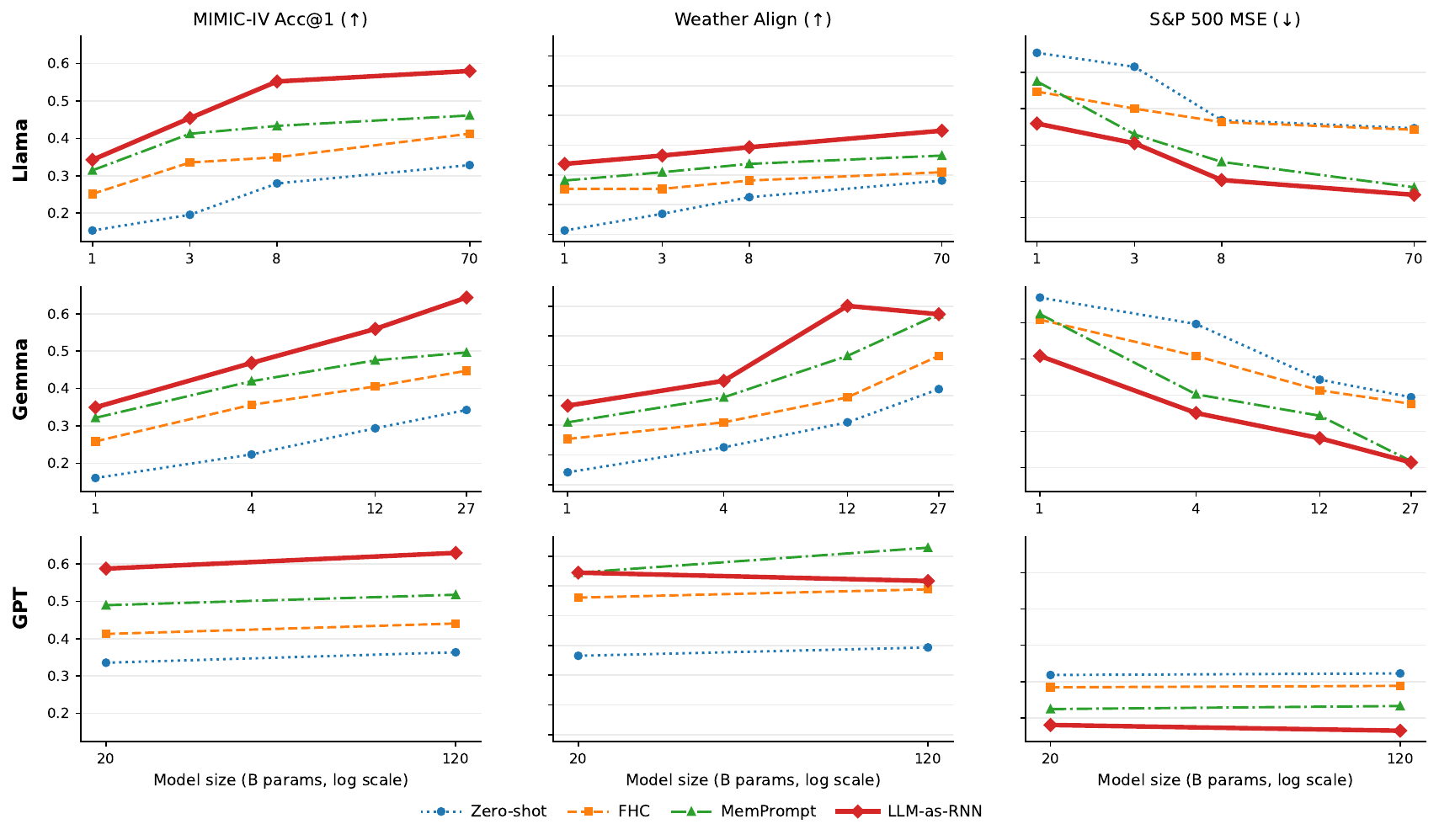}
    \caption{\textbf{Scaling across backbone families.} Performance vs.\ model size (B params, log-scale) for Zero-shot, FHC, MemPrompt, and LLM-as-RNN. Rows: Llama/Gemma/GPT backbones; columns: MIMIC Acc@1, Weather Align, S\&P 500 MSE. LLM-as-RNN yields consistent gains across sizes and families.}
    \label{fig:scaling}
\end{figure*}

\begin{figure}[t]
    \centering
    \includegraphics[width=\linewidth]{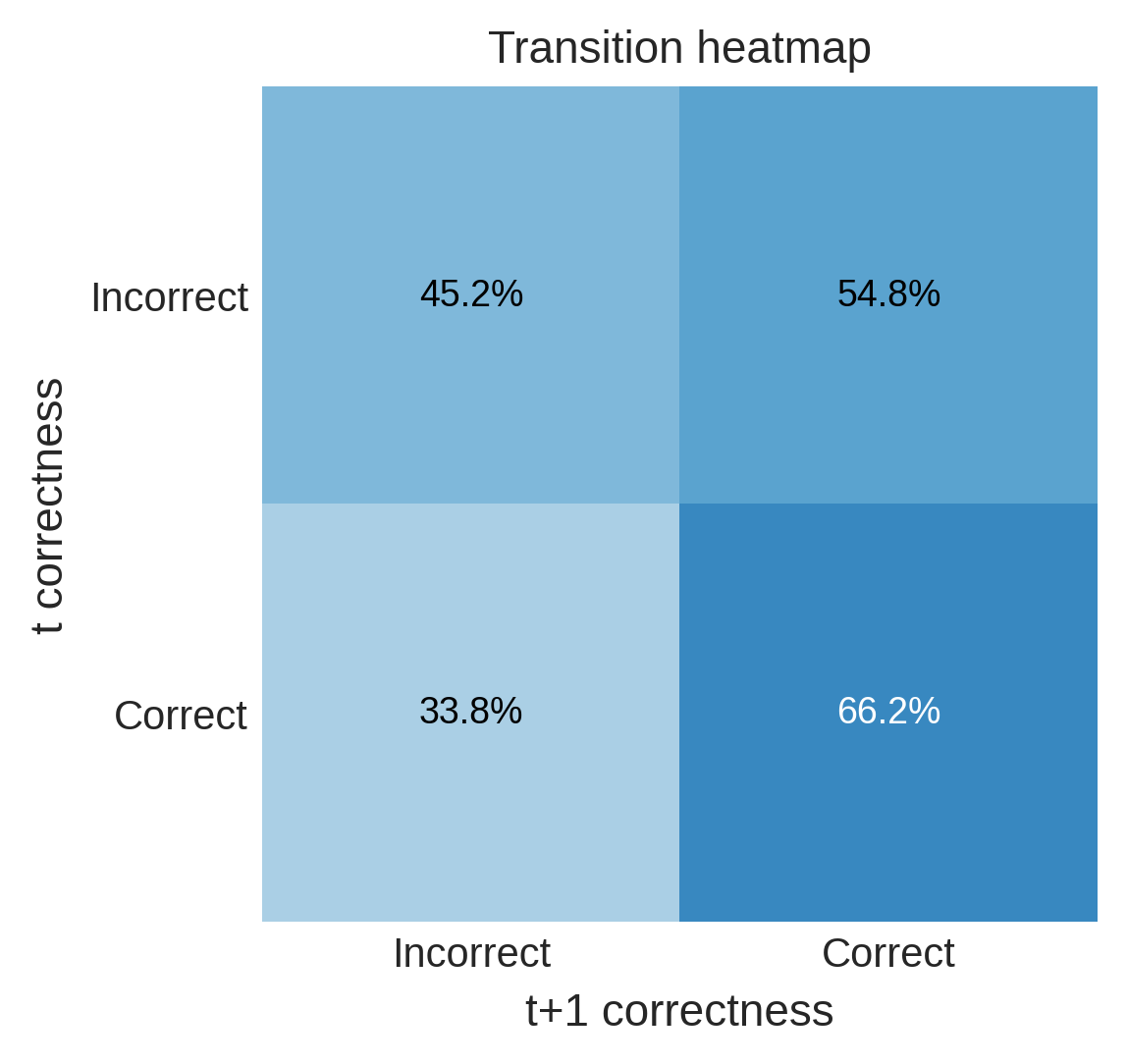}
    \caption{Transition heatmap of primary-diagnosis correctness between visits. Rows indicate correctness at time $t$ and columns indicate correctness at time $t{+}1$.}
    \label{fig:transition}
\end{figure}

\section{Prompts for MIMIC-IV datasets}
\label{sec:mimic_prompts}

In this section, we provide the full prompt templates used for the MIMIC-IV clinical diagnosis task.

% --- Configuration Start ---
% Ensure \usepackage{enumitem} is in your preamble

% 1. Configure code block style (Code wrapping & tight spacing)
\lstset{
    basicstyle=\ttfamily\footnotesize,
    breaklines=true,
    columns=fullflexible,
    keepspaces=true,
    frame=none,
    aboveskip=2pt,
    belowskip=2pt,
    escapeinside={(*@}{@*)}
}

% 2. Configure Box Style (Fixed alignment & list spacing)
\tcbset{
    promptstyle/.style={
        colframe=black!35,
        colback=black!2,
        boxrule=0.8pt,
        arc=3pt,
        fonttitle=\bfseries\small,
        fontupper=\small\raggedright, % Fixes large gaps in text
        breakable,
        left=3mm, right=3mm, top=3mm, bottom=3mm,
        before upper={ % Global list compaction within boxes
            \setlist[itemize]{nosep, leftmargin=1.2em, topsep=2pt}
        }
    }
}
% --- Configuration End ---

\subsection{Method: Initialization and Generation}

\begin{tcolorbox}[title=System Initialization, promptstyle]
You are an expert clinical AI assistant analyzing longitudinal patient records.

\textbf{Patient Information:}
\begin{itemize}
    \item Patient ID: \texttt{\{patient\_id\}}
    \item Age: \texttt{\{age\}} years
    \item Gender: \texttt{\{gender\}}
    \item Primary Conditions: \texttt{\{primary\_conditions\}}
\end{itemize}

\textbf{Evolving Clinical Summary:}
\texttt{\{evolving\_summary\}}

Your task is to provide accurate clinical predictions based on the current visit information and patient history.
\end{tcolorbox}

\begin{tcolorbox}[title=Default Initialization Value (First Visit), promptstyle]
\textbf{CLINICAL HISTORY:}
This is the patient's first visit. No previous medical history available in the system.

\textbf{CURRENT STATUS \& FOCUS AREAS:}
\begin{itemize}
    \item Establishing baseline assessment
    \item Gathering initial clinical presentation
\end{itemize}

\textbf{KEY CONSIDERATIONS FOR FUTURE VISITS:}
\begin{itemize}
    \item Baseline health status to be established after first visit
    \item Future assessments will track disease progression
\end{itemize}
\end{tcolorbox}

\begin{tcolorbox}[title=Prediction Prompt ($\mathcal{P}_{gen}$), promptstyle]
You are a clinical AI assistant predicting discharge diagnoses for the current visit.

\textbf{CURRENT VISIT SNAPSHOT:}
\begin{itemize}
    \item Chief Complaint: \texttt{\{chief\_complaint\}}
    \item Vital Signs: \texttt{\{vitals\}}
    \item Lab Results: \texttt{\{labs\}}
    \item Medications on Admission: \texttt{\{medications\}}
    \item Recent Procedures: \texttt{\{procedures\}}
    \item History of Present Illness: \texttt{\{history\_present\_illness\}}
    \item Past Medical History: \texttt{\{past\_medical\_history\}}
    \item Social History: \texttt{\{social\_history\}}
    \item Family History: \texttt{\{family\_history\}}
    \item Physical Exam: \texttt{\{physical\_exam\}}
    \item Pertinent Results: \texttt{\{pertinent\_results\}}
    \item Hospital Course: \texttt{\{hospital\_course\}}
    \item Allergies: \texttt{\{allergies\}}
\end{itemize}

\textbf{TASK:} Provide ONLY the discharge diagnosis predictions.
Return JSON with top-5 diagnoses and the primary diagnosis:
\begin{lstlisting}
{
  "top_5_diagnoses": [
    {"name": "diagnosis1"},
    {"name": "diagnosis2"},
    {"name": "diagnosis3"},
    {"name": "diagnosis4"},
    {"name": "diagnosis5"}
  ],
  "primary_diagnosis": {"name": "diagnosis1"}
}
\end{lstlisting}
\end{tcolorbox}

\subsection{Method: Reflection and Memory Update}

\begin{tcolorbox}[title=Evaluation/Reflection Prompt ($g_{eval}$), promptstyle]
You are evaluating predicted discharge diagnoses against ground truth to guide learning.

\textbf{PREDICTED OUTPUT (JSON):} \texttt{\{prediction\}}

\textbf{GROUND TRUTH DIAGNOSES:} \texttt{\{ground\_truth\_diagnoses\}}

Evaluate with SEMANTIC MATCHING (consider broader/narrower terms as correct if clinically equivalent).

Return JSON with ONLY these fields. Please respond in JSON format:
\begin{lstlisting}
{
  "diagnosis_evaluation": {
    "primary_correct": true/false,
    "any_top5_correct": true/false,
    "missed_diagnoses": ["diagnosis1", "diagnosis2"],
    "why_missed": "brief reason for misses"
  },
  "improvement_suggestions": "one concise sentence on how to improve future predictions"
}
\end{lstlisting}
\end{tcolorbox}

\begin{tcolorbox}[title=Memory Update Prompt ($\mathcal{P}_{mem}$), promptstyle]
You are updating a clinical AI system's memory to learn from prediction errors.

\textbf{CURRENT EVOLVING SUMMARY:} \texttt{\{current\_evolving\_summary\}}

\textbf{PATIENT'S ACTUAL MEDICAL HISTORY} (from all previous visits): \texttt{\{patient\_history\}}

\textbf{EVALUATION FEEDBACK} (Prediction vs Ground Truth): \texttt{\{evaluation\_feedback\}}

\textbf{VISIT CONTEXT:} \texttt{\{visit\_summary\}}

\textbf{CRITICAL INSTRUCTIONS:}
\begin{itemize}
    \item The evaluation feedback contains LLM JUDGE analysis with: what diagnoses were MISSED and WHY, what clinical signs were OVERLOOKED, and specific improvement suggestions.
    \item The PATIENT HISTORY shows the ACTUAL diagnoses and treatments across visits.
    \item Your task: Update the evolving summary to incorporate past learnings, current status, and future considerations.
\end{itemize}

\textbf{PAY SPECIAL ATTENTION TO:}
\begin{itemize}
    \item "missed\_diagnoses" - what we failed to predict
    \item "why\_missed" - root cause of our errors
    \item "improvement\_suggestions" - single concise sentence
\end{itemize}

Create a CONCISE evolving summary (MAX 200 words total) with three short sections:

1. \textbf{CLINICAL HISTORY} (2-3 sentences max): List key diagnoses from patient history with visit numbers. Note recurring conditions.

2. \textbf{FOCUS AREAS} (2-3 key points): What diagnoses were missed in this visit? Why were they missed?

3. \textbf{FUTURE CONSIDERATIONS} (1-2 key points): What to watch for based on patient's actual conditions.

CRITICAL: Keep it brief and focused. Stop after completing the JSON structure.

Provide the updated evolving summary as JSON:
\begin{lstlisting}
{
  "evolving_summary": "unified summary with three sections: CLINICAL HISTORY, CURRENT STATUS & FOCUS AREAS, and KEY CONSIDERATIONS FOR FUTURE VISITS"
}
\end{lstlisting}
\end{tcolorbox}

\subsection{Baselines}

\begin{tcolorbox}[title=Baseline: Zero-Shot, promptstyle]
You are an expert clinical AI assistant. Analyze the following patient visit and provide:

1. Top-5 diagnoses (ranked by likelihood)

2. Primary diagnosis

\textbf{Current Visit Information:}
\begin{itemize}
    \item Patient Demographics: Age \texttt{\{age\}}, Gender \texttt{\{gender\}}
    \item Chief Complaint: \texttt{\{chief\_complaint\}}
    \item Vital Signs: \texttt{\{vital\_signs\}}
    \item Lab Results: \texttt{\{lab\_results\}}
    \item Current Medications: \texttt{\{medications\}}
    \item Recent Procedures: \texttt{\{procedures\}}
\end{itemize}

\textbf{TASK:} Provide ONLY the discharge diagnosis predictions.
Return JSON with top-5 diagnoses and the primary diagnosis:
\begin{lstlisting}
{
  "top_5_diagnoses": [
    {"name": "diagnosis1"},
    {"name": "diagnosis2"},
    {"name": "diagnosis3"},
    {"name": "diagnosis4"},
    {"name": "diagnosis5"}
  ],
  "primary_diagnosis": {"name": "diagnosis1"}
}
\end{lstlisting}
\end{tcolorbox}

\begin{tcolorbox}[title=MemPrompt Summarization, promptstyle]
The following system prompt has become too long and needs compression:

\texttt{\{long\_system\_prompt\}}

\textbf{Task:} Create a compressed version that:

1. Retains all critical clinical information

2. Removes redundancy

3. Keeps the most recent and relevant updates

4. Maintains the same structure (Focus Areas, Status, Considerations)

5. Stays under 250 words

Provide the compressed prompt maintaining the same JSON structure.
\end{tcolorbox}

\begin{tcolorbox}[title=Baseline: FHC and MemPrompt, promptstyle]
You are a clinical decision support system. You have access to a patient's complete visit history with known outcomes (diagnoses and treatments).

\textbf{PATIENT VISIT HISTORY (with known outcomes):}
\texttt{\{patient\_history\}}

\textbf{CURRENT VISIT:}
\texttt{\{current\_visit\}}

\textbf{TASK:}
Based on the patient's complete history and current presentation, provide Top 5 most likely discharge diagnoses (ranked by likelihood, from most likely to least likely).

\textbf{Consider:}
\begin{itemize}
    \item Patterns from previous visits and their outcomes
    \item Disease progression and comorbidities
    \item Current clinical presentation
\end{itemize}

\textbf{TASK:} Provide ONLY the discharge diagnosis predictions.
Return JSON with top-5 diagnoses and the primary diagnosis:
\begin{lstlisting}
{
  "top_5_diagnoses": [
    {"name": "diagnosis1"},
    {"name": "diagnosis2"},
    {"name": "diagnosis3"},
    {"name": "diagnosis4"},
    {"name": "diagnosis5"}
  ],
  "primary_diagnosis": {"name": "diagnosis1"}
}
\end{lstlisting}
\end{tcolorbox}

\section{Qualitative Analysis: Memory Trace}
\label{sec:qualitative_trace}

To illustrate the recurrent inference mechanism, we present a step-by-step trace of Patient \texttt{10035631}. This example demonstrates how the \textit{Memory State} ($h_t$) evolves to correct errors and accumulate clinical context over time.

% --- Define Custom Colors for the Trace ---
\definecolor{traceinput}{HTML}{E8F0FE}  % Light Blue
\definecolor{tracepred}{HTML}{E6F4EA}   % Light Green
\definecolor{tracecrit}{HTML}{FCE8E6}   % Light Red
\definecolor{tracemem}{HTML}{FFF8E1}    % Light Yellow

% --- Define Custom Box Styles ---
\tcbset{
    tracebox/.style={
        enhanced,
        colframe=black!35,
        boxrule=0.5pt,
        arc=2pt,
        fonttitle=\bfseries\small,
        fontupper=\footnotesize,
        left=2mm, right=2mm, top=2mm, bottom=2mm,
        breakable
    },
    arrowbox/.style={
        frame hidden,
        colback=white,
        fontupper=\centering\bfseries\color{gray},
        top=0mm, bottom=0mm
    }
}

% ================= VISIT 1 =================
\subsection*{Step 1: Initialization and Initial Feedback (V1)}
The patient presents with Leukemia. The model predicts the primary condition correctly but misses secondary electrolyte abnormalities. The memory is updated to watch for these in the future.

\begin{tcolorbox}[tracebox, title=Input ($x_1$) \& Prediction ($\hat{y}_1$), colback=traceinput]
\textbf{Chief Complaint:} Leukemia \\
\textbf{Prediction:} Acute Myeloid Leukemia (AML) in remission. \\
\textbf{Ground Truth:} Leukemia, Aspergillosis, Pancytopenia, \textbf{Hypokalemia}.
\end{tcolorbox}

\begin{tcolorbox}[tracebox, title=Reflection/Feedback ($s_1$), colback=tracecrit]
\textbf{Error Analysis:} Primary diagnosis correct. However, electrolyte abnormalities like \textbf{hypokalemia} were missed. \\
\textbf{Guidance:} Include common metabolic/electrolyte abnormalities in the differential for hematologic malignancy patients.
\end{tcolorbox}

\begin{tcolorbox}[tracebox, title=Memory Update ($h_0 \to h_1$), colback=tracemem]
\textbf{Previous Memory:} [Empty/Default Initialization] \\
\rule{\linewidth}{0.4pt} \\
\textbf{Updated Memory ($h_1$):} \\
\textit{Clinical History:} Leukemia (V1), Aspergillosis (V1), Hypokalemia (V1). \\
\textit{Future Focus:} \textbf{Watch for metabolic and electrolyte abnormalities} (specifically Potassium) in hematologic patients.
\end{tcolorbox}

\begin{center} \textit{... Visits 2, 3, and 4 processed (Memory accumulates breast cancer, pneumonia) ...} \end{center}

% ================= VISIT 5 =================
\subsection*{Step 2: Error Correction via Recurrence (V5)}
In Visit 5, the model falsely predicts "Remission" when the patient has "Relapsed". The reflection module catches this, and the memory explicitly encodes this correction to prevent future complacency.

\begin{tcolorbox}[tracebox, title=Input ($x_5$) \& Prediction ($\hat{y}_5$), colback=traceinput]
\textbf{Chief Complaint:} Admission for cycle 1 of Dacogen (chemotherapy). \\
\textbf{Prediction:} Leukemia (AML) \textbf{in remission}. \\
\textbf{Ground Truth:} \textbf{Relapsed} AML.
\end{tcolorbox}

\begin{tcolorbox}[tracebox, title=Reflection/Feedback ($s_5$), colback=tracecrit]
\textbf{Error Analysis:} Primary diagnosis incorrectly identified AML as 'in remission' when patient actually had relapsed disease. \\
\textbf{Guidance:} Prioritize active disease states over remission status when clinical evidence (Dacogen) suggests relapse.
\end{tcolorbox}

\begin{tcolorbox}[tracebox, title=Memory Update ($h_4 \to h_5$), colback=tracemem]
\textbf{Previous Focus:} Watch for pulmonary complications, GVHD. \\
\rule{\linewidth}{0.4pt} \\
\textbf{Updated Focus ($h_5$):} \\
\textit{Missed Diagnosis Note:} Primary diagnosis incorrectly identified AML as 'in remission'. \\
\textit{Improvement Strategy:} \textbf{Prioritize active disease states} over remission status. Ensure most acute condition is selected.
\end{tcolorbox}

% ================= FINAL STATE =================
\subsection*{Step 3: Long-Term Memory Retention (V8)}
By the end of the sequence, the memory state ($h_T$) has become a comprehensive summary of the patient's complex trajectory, far exceeding the context window of a standard zero-shot prompt.

\begin{tcolorbox}[title=Final Memory State ($h_8$), promptstyle]
\textbf{CLINICAL HISTORY:}
\begin{itemize}[nosep, leftmargin=1em]
    \item \textbf{Oncology:} Leukemia (V1, V8), Invasive Ductal Carcinoma (V2), Relapsed AML (V5), Multiple Myeloma (V6).
    \item \textbf{Infectious:} Aspergillosis (V1), Sepsis (V6), Neutropenic fever (V7).
    \item \textbf{Comorbidities:} Hypokalemia (V1), Acute Kidney Failure (V6), GVHD (V3), Orthostatic Hypotension (V4).
\end{itemize}

\textbf{CURRENT STATUS \& FOCUS AREAS:}
\begin{itemize}[nosep, leftmargin=1em]
    \item \textbf{Learned Patterns:} Patient has recurring Heart Failure (V1, V3, V5).
    \item \textbf{Correction History:} Watch for specific organisms in bacteremia (missed in V7); distinguish between active relapse vs remission.
\end{itemize}

\textbf{KEY CONSIDERATIONS FOR FUTURE VISITS:}
\begin{itemize}[nosep, leftmargin=1em]
    \item Watch for pulmonary complications in hematologic malignancy.
    \item Monitor electrolyte levels (Potassium) carefully.
    \item Distinguish between Leukemia vs Myeloma presentation.
\end{itemize}
\end{tcolorbox}

\section{Error Analysis}
\label{error_analysis}

\subsection{Non-parsable outputs.}
With smaller backbones or higher sampling randomness, generations more often violate strict JSON-only constraints (e.g., emitting extra natural-language text, markdown code fences, trailing commas, or missing braces). These formatting failures break automatic parsing and can halt the recurrent pipeline, since both the evaluator and the memory update step depend on structured fields to propagate feedback across timesteps.

\subsection{Overlong generations causing truncation.}
Prediction, critique, or memory-update outputs can exceed \texttt{max\_tokens} or the memory budget $\lambda$, leading to truncation. This is particularly damaging when truncation cuts off JSON closures (making outputs non-parsable) or removes critical supervision signals such as \texttt{missed\_diagnoses} and \texttt{why\_missed}. In a recurrent setting, losing these fields not only degrades the current timestep but also weakens the next memory update, compounding error over long horizons.

\subsection{Noisy or biased feedback.}
When ground truth labels are unavailable and feedback is generated by an LLM judge, the critique can be noisy, inconsistent, or biased (e.g., over-penalizing acceptable synonyms, missing clinically equivalent diagnoses, or providing spurious rationales). Because the memory update treats this feedback as a ``semantic gradient,'' systematic judge errors can cause the state to internalize incorrect lessons, inducing memory drift and potentially reinforcing mistakes across subsequent timesteps.

\end{document}